\documentclass[10pt,twocolumn,letterpaper]{article}

\usepackage{cvpr}
\usepackage{times}
\usepackage{epsfig}
\usepackage{graphicx}
\usepackage{amsmath}
\usepackage{amssymb}

\usepackage{enumitem}
\usepackage{amsmath}
\usepackage{algorithm}
\usepackage[noend]{algpseudocode}
\usepackage[font=small,labelfont=bf]{caption}
\usepackage{graphicx}
\usepackage{subfig}

\usepackage[utf8]{inputenc} 
\usepackage[T1]{fontenc}    
\usepackage{url}            
\usepackage{booktabs}       
\usepackage{amsfonts}       
\usepackage{nicefrac}       
\usepackage{microtype}      
\usepackage{soul}
\usepackage{setspace}
\usepackage{makecell}
\usepackage{longtable}
\usepackage{placeins}
\usepackage[table]{xcolor}

\definecolor{Gray}{gray}{0.90}
\definecolor{HeaderGray}{gray}{0.65}
\newcolumntype{g}{>{\columncolor{Gray}}r}

\usepackage[pagebackref=true,breaklinks=true,letterpaper=true,colorlinks,bookmarks=false]{hyperref}

\cvprfinalcopy 

\def\cvprPaperID{8651} 

\ifcvprfinal\fi
\begin{document}

\title{Latent Replay for Real-Time Continual Learning}

\author{Lorenzo Pellegrini, Gabriele Graffieti, Vincenzo Lomonaco, Davide Maltoni\\
University of Bologna\\
Via dell'Università, 50, Cesena, Italy\\
{\tt\small \{l.pellegrini, gabriele.graffieti, vincenzo.lomonaco, davide.maltoni\}@unibo.it}\\
{\tt\small\url{http://bit.ly/latent-replay}}
}

\renewcommand{\baselinestretch}{1.0}
\setstretch{1.0}
\renewcommand\theadfont{\small}

\maketitle
\newcommand{\red}[1]{\textcolor{red}{#1}}

\begin{abstract}
\noindent Training deep neural networks at the edge on light computational devices, embedded systems and robotic platforms is nowadays very challenging. Continual learning techniques, where complex models are incrementally trained on small batches of new data, can make the learning problem tractable even for CPU-only embedded devices enabling remarkable levels of adaptiveness and autonomy. However, a number of practical problems need to be solved: catastrophic forgetting before anything else. In this paper we introduce an original technique named ``Latent Replay'' where, instead of storing a portion of past data in the input space, we store activations volumes at some intermediate layer. This can significantly reduce the computation and storage required by native rehearsal. To keep the representation stable and the stored activations valid we propose to slow-down learning at all the layers below the latent replay one, leaving the layers above free to learn at full pace. In our experiments we show that Latent Replay, combined with existing continual learning techniques, achieves state-of-the-art performance on complex video benchmarks such as CORe50 NICv2 (with nearly 400 small and highly non-i.i.d. batches) and OpenLORIS. Finally, we demonstrate the feasibility of nearly real-time continual learning on the edge through the deployment of the proposed technique on a smartphone device.
\end{abstract}

\section{Introduction}

Training on the edge (e.g., on light computing devices such as smartphones, smart cameras, embedded systems and robotic platforms) is highly desirable in several applications where privacy, lack of network connection and fast adaptation are real constraints. While some steps in this direction have been recently moved \cite{Li2019i}, training on the edge often remains unfeasible. In fact, given the high demand in terms of memory and computation, most machine learning models nowadays are trained on powerful multi-GPUs servers, and only frozen models are deployed to edge devices for inference.

\begin{figure}[h]
\centering
\includegraphics[width=\columnwidth]{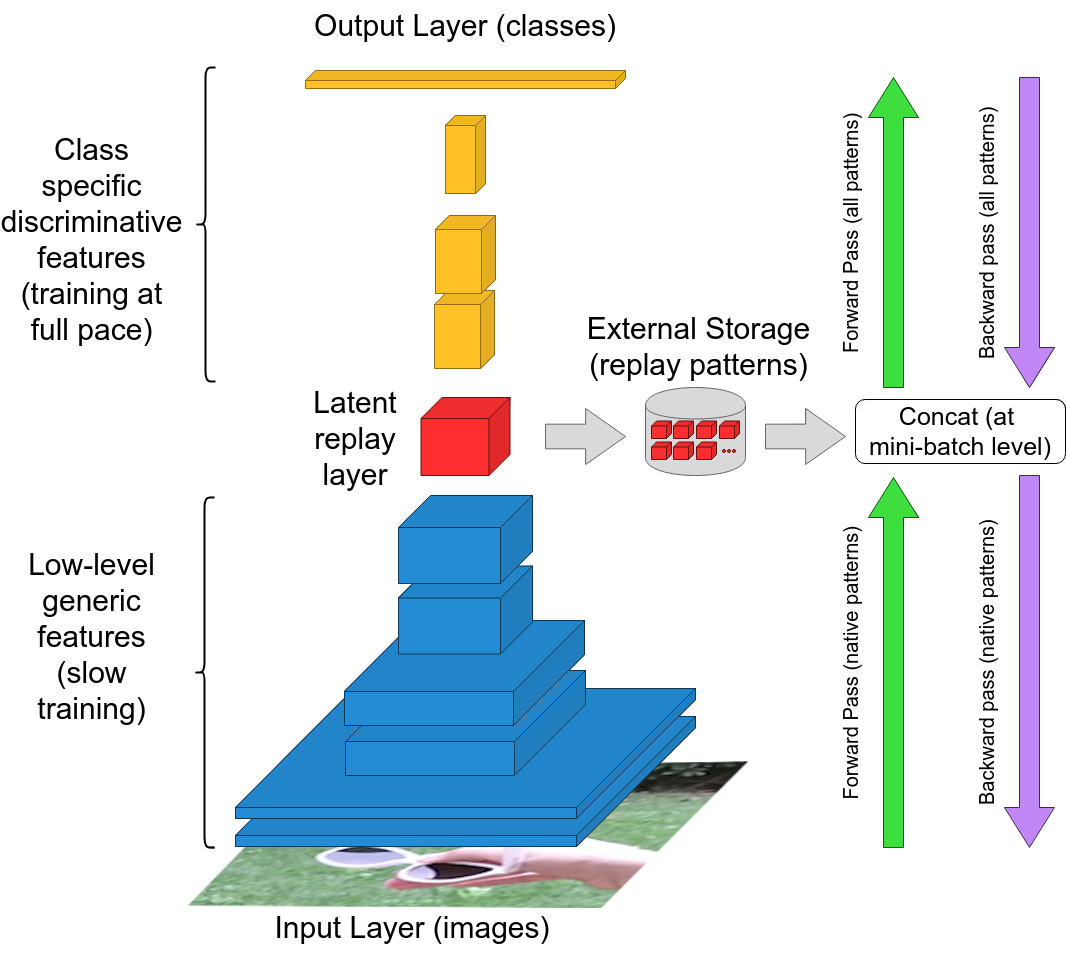}
\caption{Architectural diagram of Latent Replay.}
\label{fig:latent_replay}
\end{figure}

\begin{figure*}[t]
\centering
\includegraphics[width=\textwidth]{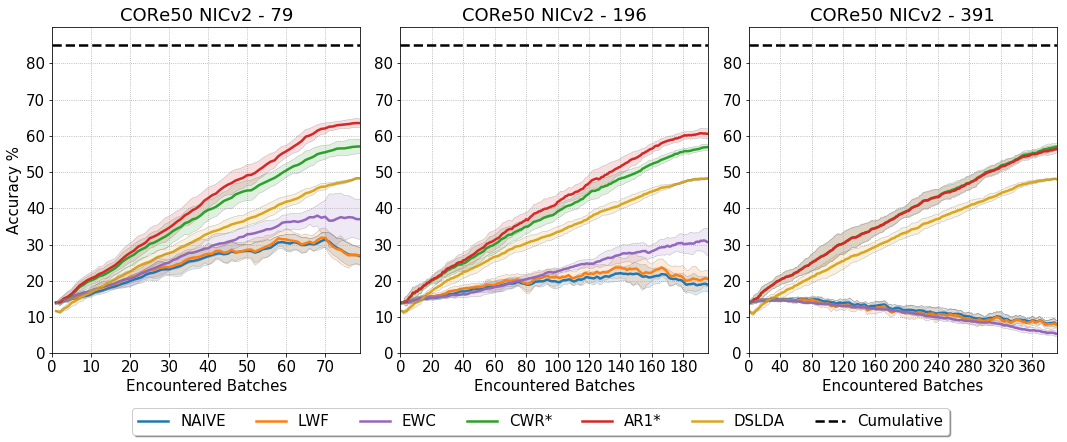}
\caption{Continual learning accuracy along the incremental training batches over NICv2 – 79, NICv2 – 196 and NICv2 – 391 as presented in \cite{Lomonaco2019}. None of the method compared uses rehearsal. Naive refers to a simple approach where the model is tuned along the batches and the only protection against forgetting is early stopping. LWF and EWC are well-known methods for CL (see \cite{Li2016,Kirkpatrick2016}). CWR* and AR1* are discussed in the main text. DSLDA is a recently proposed streaming continual learning approach \cite{Hayes2019}. The black dashed line denotes the ``upper bound'' accuracy achieved by the Cumulative approach, that is a full single training on the entire dataset.}
\label{fig:fg_results}
\end{figure*}

Furthermore, in some applications (e.g., robotic vision, see Fig. \ref{fig:robot}), training a deep model from scratch as soon as new data becomes available is prohibitive in terms of storage / computation even if performed server side. Continual Learning (CL), that is the ability of continually training existing models using only new data, is gaining a lot of attention and several solutions have been recently proposed to deal with the daunting issue of catastrophic forgetting (i.e., as the model learns new concepts and skills, it tends to forget the old ones) \cite{McCloskey1989}. Recent surveys \cite{Parisi2019, Lesort2019} provide an overview of the CL field. In principle, CL approaches could be exploited not only to control forgetting but also to reduce the training complexity. 

In this paper we focus on real-time CL and prove that continual training with small batches can be compatible with the limited computing power made available by CPU-only embedded devices and robotic platforms.

In \cite{Lomonaco2019} it was shown that some CL approaches can effectively learn to recognize objects (on the CORe50 dataset \cite{pmlr-v78-lomonaco17a}) even when fed with fine-grained incremental batches. CORe50 NICv2 \cite{Lomonaco2019} is a continual learning benchmark where objects from 50 different classes have to be learned incrementally. What makes this benchmark challenging is that classes are discovered a little at a time and the training batches are small and non i.i.d. In particular, in NICv2 - 391 each training batch includes only 300 frames extracted from a short video (15 seconds at 20 fps) of a single object slowly moving in front of the camera: hence, patterns within each batch are highly correlated. Despite these nuisances, in \cite{Lomonaco2019} the approaches denoted as CWR* and AR1* proved to be able to learn continually even in absence of replay mechanisms, that is, the periodic refresh of old examples maintained in an external memory. Under this challenging setting both CWR* and AR1* performed significantly better than well known techniques such as LWF \cite{Li2016} and EWC \cite{Kirkpatrick2016}. While these results are encouraging:

\begin{itemize}
\item the accuracy gap w.r.t. the cumulative approach (a sort of upper bound obtained by training the model on the entire training set) remains quite relevant (about 20\%).
\item in NICv2 - 391, the most challenging setup, AR1* was not able to effectively adapt the representation layers during continual learning.
\end{itemize}

The aim of this work is to \emph{reduce as much as possible the gap w.r.t. the cumulative upper bound} and, at the same time, to \emph{provide an efficient implementation strategy of CL approaches to enable nearly real-time training on the edge}.\\

To this purpose we first show that a small amount of pattern replay is sufficient to significantly improve accuracy on NICv2 - 391 (Section \ref{sec:native_rehe}). However, even if in the CORe50 setting the extra memory required by replay is not an issue (we store only 30 patterns for each of the 50 classes), a constant refresh significantly increases the required amount of computation because of the extra forward and backward steps and this makes the resulting training too resource demanding for real-time applications. Therefore, we propose a \emph{``Latent Replay''} approach (Section \ref{sec:latent_replay}) where old data are injected at some intermediate layer selected according to the desired accuracy-efficiency trade-off. 

In Section \ref{sec:train_latent_replay} we compare our approach with several continual learning algorithms and show its advantages reaching state-of-the-art performances on two different benchmarks: CORe50 and OpenLORIS. Finally, in order to demonstrate the practical applicability of the proposed approach, in Section \ref{sec:mobile} we discuss the implementation of a continual learning application for Android smartphones that, starting from a pre-trained MobileNetV1 model with 10 classes, can incrementally learn (in near real-time) new classes and/or new objects of existing classes.

\section{Related Works}
\label{sec:related_works}

While several works in the continual learning literature focus on Multiple independent Tasks (MT) scenario, in many practical applications such as robotic vision, a Single Incremental Task (SIT) scenario is more appropriate \cite{maltoni2019}. In particular, a robot should be able to incrementally improve its object recognition capabilities while being exposed to new instances of both known and completely new classes (denoted as NIC setting - New Instances and Classes). CORe50 NICv2 benchmark specifically addresses this problem \cite{Lomonaco2019}. Other datasets have been released to study continual learning for robotic vision (e.g., iCub-transformation \cite{Pasquale2019a}, OpenLORIS \cite{openloris}) but no NIC benchmarks have been yet defined for them. ImageNet-1K \cite{Li2009a} and CIFAR-100 \cite{Krizhevsky2009} have also been used to evaluate continual learning techniques, but these datasets do not fit well the object recognition task because of the lack of multiple videos of the same objects taken under different poses, lighting and backgrounds.

In \cite{Lomonaco2019}, two approaches denoted as CWR* and AR1* have been evaluated on CORe50 NICv2 (see Figure \ref{fig:fg_results}): in CWR* the last fully connected layer is implemented as a double memory, and simple initialization and fusions steps are performed before and after each training batch to synchronize the two memories. However, after the first training batch, CWR* freezes all the layers except the last one, thus losing the benefits of a continual adaptation of the underlying representation. AR1* extends CWR* by enabling end-to-end continual training throughout the entire network; to this purpose the Synaptic Intelligence \cite{Zenke2017} regularization approach (similar to Online-EWC \cite{Schwarz2018}) is adopted to constrain the change of critical weights.

\begin{figure}[t]
\centering
\includegraphics[width=0.95\columnwidth]{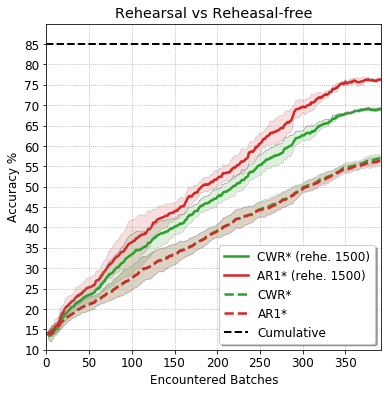}
\caption{Comparison of CWR* and AR1* on CORe50 NICv2 – 391 with and without rehearsal ($RM_{size} = 1500$). Each experiment was averaged on 5 runs with different batch ordering: colored areas represent the standard deviation of each curve. The black dashed line denotes the reference accuracy of the cumulative upper bound.}
\label{fig:rehe_exps}
\end{figure}

Patterns replay, which is central in the proposed approach, proved to be an effective approach to contrast forgetting in continual learning scenarios \cite{Rebuffi2017, Lopez-paz2017, Rolnick2018, pomponi}. In fact, periodically replaying some representative patterns from old data helps the model to retain important information of past tasks / classes while learning new concepts. iCaRL \cite{Rebuffi2017} uses well-designed entry / exit criteria (denoted as herding) to maintain a class-balanced set of exemplars that maximize representativeness. A comparison between the proposed technique and iCaRL is reported in Section \ref{subsec:comparison}. Generative Replay (also known as \emph{``Pseudo-rehearsal''} \cite{Robins1995}), where surrogates of past data are generated without explicitly storing native patterns, looks very appealing because of the storage saving; however, most of the proposed approaches to date do not allow on-line generation of effective replay patterns.

\begin{figure*}[t]
\centering
\includegraphics[width=\textwidth]{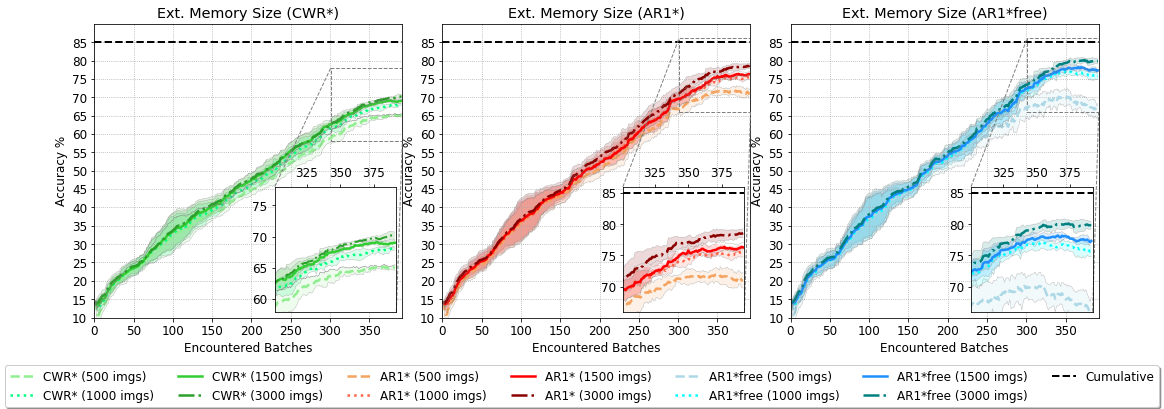}
\caption{Comparison of CWR*, AR1* and AR1*free on CORe50 NICv2 – 391 with different external memory sizes ($RM_{size} = 500, 1000, 1500$ and $3000$ patterns).}
\label{fig:mem_size}
\end{figure*}

Another class of relevant techniques for our study are the so called \emph{streaming} continual learning approaches \cite{Hayes2018a}, where a model can be incrementally trained with a single pattern at a time. Even if in a robotic vision scenario learning from single frames does not appear necessary (in fact, using short videos of single objects can be more
efficient and looks more biologically plausible) efficient streaming learning techniques can be effortlessly applied to NIC setting. Deep Streaming Linear Discriminant Analysis (DSLDA) was recently proposed \cite{Hayes2019} where an online extension of the LDA classifier works on the top of a fixed deep learning feature extractor. This approach, which achieved state-of-the-art accuracy on (partitioned) ImageNet-1K and CORe50 (10 classes version) was run on NICv2 and compared with other techniques in Figure \ref{fig:fg_results}. 

Finally, training on the edge was recently addressed in \cite{Li2019i}, where an object detection model was incrementally trained based on LWF and pattern replay. While training is not actually real-time (it requires a few minutes on Nvidia Jetson TX2 board) and only few large continual training batches are presented to the model, the detection problem approached in \cite{Li2019i} is more difficult than the classification problem here considered and therefore we cannot make a direct comparison.

\section{Native Rehearsal}
\label{sec:native_rehe}

In \cite{maltoni2019} it was shown that a very simple rehearsal implementation (hereafter denoted as \emph{native rehearsal}), where for every training batch a random subset of the batch patterns is added to the external storage to replace a (equally random) subset of the external memory, is not less effective than more sophisticated approaches such as iCaRL.  Therefore, in this study we opted for simplicity and started by expanding CWR* and AR1* with the trivial rehearsal approach summarized in Algorithm \ref{algo:rm}. In Figure \ref{fig:rehe_exps} we compare the learning trend of CWR* and AR1* of a MobileNetV1\footnote{The network was pre-trained on ImageNet-1k.} trained with and without rehearsal on CORe50 NICv2 – 391. We use the same protocol and hyper-parameters introduced in \cite{Lomonaco2019} and a rehearsal memory of 1,500 patterns. It is well evident that even a moderate external memory (about 1.27\% of the total training set) is very effective to improve the accuracy of both approaches and to reduce the gap with the cumulative upper bound that for this model is $\sim$85\%.

\begin{algorithm}[h]
\captionsetup{font=small}
\begin{algorithmic}[1]
\vspace{0.1cm}
\footnotesize
\State $RM=\varnothing$
\State $RM_{size} = $ number of patterns to be stored in $RM$
\State \textbf{for each} $\text{training batch } B_i$:
\State \ \ \ \ train the model on shuffled $B_i \cup RM$
\State \ \ \ \ $h = \dfrac{RM_{size}}{i}$
\State \ \ \ \ $R_{add} =$ random sampling $h$ patterns from $B_i$
\State \ \ \ \ $R_{replace} = 
    \begin{cases}
      \varnothing & \text{if $i==1$}\\
      \text{random sample } h \text{ patterns from } RM & \text{otherwise}
    \end{cases}$
\State \ \ \ \ $RM = (RM - R_{replace}) \cup R_{add}$
\end{algorithmic}
\caption{Pseudocode explaining how the external memory
$RM$ is populated across the training batches. Note that the amount
$h$ of patterns to add progressively decreases to maintain a nearly
balanced contribution from the different training batches, but no
constraints are enforced to achieve a class-balancing.}
\label{algo:rm}
\end{algorithm}

To understand the influence of the external memory size we repeated the experiment with different $RM_{size}$ values: 500, 1,000, 1,500, 3,000. Since rehearsal itself protects the model from forgetting we also run AR1* (where important weights of lower layers are protected from forgetting by using Synaptic Intelligence \cite{Zenke2017} regularization) without Synaptic Intelligence protection, that is lower layers weights are left totally unconstrained; in the following we denote this approach as AR1*free. The results are shown in Figure \ref{fig:mem_size}: it is worth noting that increasing the rehearsal memory leads to better accuracy for all the algorithms, but the gap between 1500 and 3000 is not large and we believe 1500 is a good trade-off for this dataset. AR1*free works slightly better that AR1* when a sufficient number of rehearsal patterns are provided but, as expected, accuracy is worse with light (i.e. $RM_{size} = 500$) or no rehearsal.

It is worth noting that the best combination in Figure \ref{fig:mem_size} (AR1*free with 3000 patterns) is only 5\% worse than the cumulative upper bound and a better parametrization and exploitation of the rehearsal memory could further reduce this gap.

\section{Latent Replay}
\label{sec:latent_replay}

In deep neural networks the layers close to the input (often denoted as representation layers) usually perform low-level feature extraction and, after a proper pre-training on a large dataset (e.g., ImageNet), their weights are quite stable and reusable across applications. On the other hand, higher layers tend to extract class-specific discriminant features and their tuning is often important to maximize accuracy.

With latent replay (see Figure \ref{fig:latent_replay}) we denote an approach where, instead of maintaining copies of input patterns in the external memory in the form of raw data, we store the activations volumes at a given layer (denoted as \emph{Latent Replay layer}). To keep the representation stable and the stored activations valid we propose to slow-down the learning at all the layers below the latent replay one and to leave the layers above free to learn at full pace. In the limit case where lower layers are completely frozen (i.e., slow-down to 0) latent replay is functionally equivalent to rehearsal from the input, but achieves a computational and storage saving thanks to the smaller fraction of patterns that need to flow forward and backward across the entire network and the typical information compression that networks perform at higher layers.

In the general case where the representation layers are not completely frozen, the activations stored in the external memory suffer from an aging effect (i.e., as the time passes they tend to increasingly deviate from the activations that the same pattern would produce if feed-forwarded from the input layer). However, if the training of these layers is sufficiently slow, the aging effect is not disruptive since the external memory has enough time to be rejuvenated with fresh patterns. When latent replay is implemented with mini-batch SGD training: \emph{(i)} in the forward step, a concatenation is performed at the replay layer (on the mini-batch dimension) to join patterns coming from the input layer with activations coming from the external storage; \emph{(ii)} the backward step is stopped just before the replay layer for the replay patterns.

\section{Experiments and Results}
\label{sec:train_latent_replay}

Hereafter, while the proposed latent replay approach is architecture agnostic, we discuss its specific design with several continual learning algorithms CWR*, AR1*, AR1*free and LWF over a MobileNet \cite{Howard2012} pre-trained on ImageNet-1K. We compare our latent replay approach with other state-of-the-art techniques on CORe50 and OpenLORIS.

\subsection{Experiments on CORe50 NICv2 - 391}
\label{sec:core50}

For the CORe50 experiments:
\begin{itemize}
\item we use a MobileNetV1 and focus on CWR*, AR1* which have been already proved to be competitive on this benchmark.
\item for all the methods the output layer (\texttt{fc7}) must be implemented as a double memory with proper (pre)initialization and (post)fusion for each training batch (for details see CWR* pseudocode in Algorithm 2 of \cite{Lomonaco2019});
\item for CWR* the latent replay layer is the second-last layer (i.e., \texttt{pool6});
\item for AR1* and AR1*free the latent replay layer can be pushed down and selected according to the accuracy-efficiency trade-off discussed below;
\item for AR1*free the Synaptic Intelligence regularization is switched off.
\end{itemize}

\begin{figure}[h]
\centering
\includegraphics[width=\columnwidth]{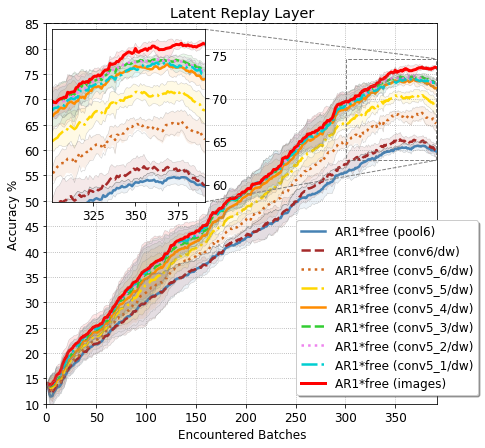}
\caption{AR1*free with latent replay ($RM_{size}=1500$) for different choices of the latent replay layer. Setting the replay layer at the \texttt{pool6} layer makes AR1*free equivalent to CWR*. Setting the replay layer at the ``images'' layer corresponds to native rehearsal (same curve of Figure \ref{fig:mem_size} for AR1*free and 1500 patterns). The 
saturation effect which characterizes the last training batches is due to the data distribution in NICv2 – 391 (see \cite{Lomonaco2019}): in particular, the lack of new instances for some classes (that already introduced all their data) slows-down the accuracy trend and intensifies the effect of activations aging.}
\label{fig:latent_layer_diff}
\end{figure}

To simplify the network design and training we keep the proportion of original and replay patterns fixed: for example, if the training batches contain 300 patterns and the external memory 1500 patterns, in a mini-batch of size 128 we concatenate 21 ($128\times300 / 1800$) original patterns (of the current batch) with 107 ($128\times1500/1800$) replay patterns. In this case only 21 patterns (over 128) need to travel across the blue layers in Figure \ref{fig:latent_replay}.


\begin{table*}[th]
  \caption{Computation, storage, and accuracy trade-off with Latent Replay at different layers of a MobileNetV1 ConvNet trained continually on NICv2 – 391 with $RM_{size} = 1500$. Computation and pattern size can be easily extrapolated from Table \ref{tab:model_arch} in the appendix where the network architecture is exploded by reporting neurons, connections and weights at each layer.}
  \label{tab:tradeoff}
  \centering
  \small
  \renewcommand{\arraystretch}{1.05}
  \setlength\tabcolsep{0.4cm}
  \begin{tabular}{p{2cm}cccc}
    \toprule
    \thead[l]{Layer} & \thead{Computation \% \\ vs Native Rehearsal} & \thead{Pattern Size} & \thead{Final Accuracy} \% & \thead{$\Delta$ Accuracy \% \\ vs Native Rehearsal}\\
    \midrule
    Images & 100.00\% & 49152 & 77.30\% & 0.00\% \\
    conv5\_1/dw & 59.261\% & 32768 & 72.82\% &-4.49\%\\
    conv5\_2/dw & 50.101\% & 32768 & 73.21\% &-4.10\%\\
    conv5\_3/dw & 40.941\% & 32768 & 73.22\% & -4.09\% \\
    \textbf{conv5\_4/dw} & \textbf{31.781\%} & \textbf{32768} & \textbf{72.24\%} & \textbf{-5.07\%} \\
    conv5\_5/dw & 22.621\% & 32768 & 68.59\% & -8.71\% \\
    conv5\_6/dw & 13.592\% & 8192 & 65.24\% & -12.06\% \\
    conv6/dw & 9.012\% & 16384 & 59.89\% & -17.42\% \\
    pool6 & 0.027\% & 1024 & 59.76\% & -17.55\% \\
    \bottomrule
  \end{tabular}
\end{table*}

Concerning the learning slow-down in the representation layers we found that an effective (and efficient) strategy is blocking the weight changes after the first batch (i.e., learning rate set to 0), while leaving the batch normalization moments free to adapt to the statistics of the input patterns across all the batches. Batch Normalization (BN) \cite{loffe2015} is widely used in modern deep neural networks (including MobileNets) to control internal covariate shift thus making learning faster and more robust.
Replacing BN with Batch Renormalization (BRN) \cite{loffe2017} was proved to be a very important step for effective continual learning with fine-grained non-i.d.d. batches \cite{Lomonaco2019}, so in the MobileNetV1 here adopted BN layers have been replaced with BRN layers. In the context of latent replay, if we leave the BRN moments free to adapt, the activations stored in the external memory suffer the aging effect described in Section \ref{sec:latent_replay}. However, we experimentally verified that, upon proper setting of the global moment mobile windows (more details are provided in the Appendix \ref{app:hyper}), the accuracy drop due to the aging effect is quite limited and in any case the final accuracy is higher w.r.t. the case where BRN moments in the representation layers are frozen. On the computational side, blocking the weight changes in the representation layers allows to skip the backward pass in the lower part of the network also for native patterns, since updating the BRN moments only relies on the forward pass. 

In Figure \ref{fig:latent_layer_diff} we report the accuracy of AR1*free with latent replay ($RM_{size}=1500$) for different choices of the rehearsal layer (reported between parenthesis). As expected, when the replay layer is pushed down the corresponding accuracy increases, proving that a continual tuning of the representation layers is important. However, after \texttt{conv5\_4/dw} there is a sort of saturation and the model accuracy is no longer improving. The residual gap ($\sim$4\%) with respect to native rehearsal is not due to the weights freezing of the lower part of the network but to the aging effect introduced above. This can be simply proved by implementing an ``intermediate'' approach that always feeds the replay pattern from the input and stops the backward at \texttt{conv5\_4}: such intermediate approach achieved an accuracy at the end of the training very close to the native rehearsal. We believe that the accuracy drop due to the aging effect can be further reduced with better tuning of BNR hyper-parameters and/or with the introduction of a scheduling policy making the global moment mobile windows wider as the continual learning progresses (i.e., more plasticity in the early stages and more stability later); however, such fine optimization is application specific and beyond the scope of this study.

\begin{table*}[t]
  \caption{Summary of the computation, memory, and accuracy trade-off for each strategy. Memory overhead include both the data used for replay purposes as well as additional trainable parameters needed for continual learning. Each metric is averaged across 10 runs.}
  \label{tab:comparison}
  \centering
  \small
  \renewcommand{\arraystretch}{1.0}
  \setlength\tabcolsep{0.4cm}
  \begin{tabular}{p{3.2cm}cccc}
    \toprule
    \thead[l]{Strategy} & \thead{Run Time \\ (Minutes)} & \thead{Mem. Overhead \\(Data + Params, MB)} & \thead{Final Accuracy \%} & \thead{$\Delta$ Acc. \% \\ vs Cumulative}\\
    \midrule
    CWR* & 21.4 & 0 + 0.2 & 56.99\% & -28.27\% \\
    AR1*free (pool6) & 23.7 & 5.8 + 12.4 & 59.75\% & -25.51\% \\
    AR1* & 39.9 & 0 + 12.4 & 56.32\% &-28.94\%\\
    \textbf{AR1*free (conv5\_4/dw)} & \textbf{41.2} & \textbf{48 + 0} & \textbf{72.23\%} & \textbf{-13.03\%} \\
    DSLDA & 79.1 & 0 + 0.2 & 48.02\% & -37.24\% \\
    iCaRL & 20185.0 & 375 + 0 & 15.65\% & -69.61\% \\
    \bottomrule
  \end{tabular}
\end{table*}

\subsubsection{On the Computation, Storage and Accuracy Trade-off}
\label{subsec:tradeoff}

To better evaluate the latent replay w.r.t. native rehearsal we report in Table \ref{tab:tradeoff} the relevant dimensions: \emph{(i)} computation refers to the percentage cost in terms of ops of a partial forward (from the latent replay layer on) relative to a full forward step from the input layer; \emph{(ii)} pattern size is the dimensionality of the pattern to be stored in the external memory (considering that we are using a MobileNetV1 with 128 $\times$ 128 $\times$ 3 inputs to match CORe50 image size); \emph{(iii)} accuracy and $\Delta$ accuracy quantify the absolute accuracy at the end of the training and the gap with respect to a native rehearsal, respectively. 

For example, \texttt{conv5\_4/dw} exhibits an interesting trade-off because the computation is about 32\% of the native rehearsal one, the storage is reduced to 66\% (more on this point in subsection \ref{subsec:active_storage}) and the accuracy drop is mild (5.07\%). CWR* (i.e. AR1* with latent replay layer $\equiv$ \texttt{pool6}) has a really negligible computational cost (0.027\%) with respect to native rehearsal and still provides and accuracy improvement of $\sim$4\% w.r.t. the non-rehearsal case ($\sim$60\% vs $\sim$56\% as it is possible to see from Figure \ref{fig:latent_layer_diff} and Figure \ref{fig:mem_size}, respectively).

\begin{figure}[h]
\centering
\includegraphics[width=\columnwidth]{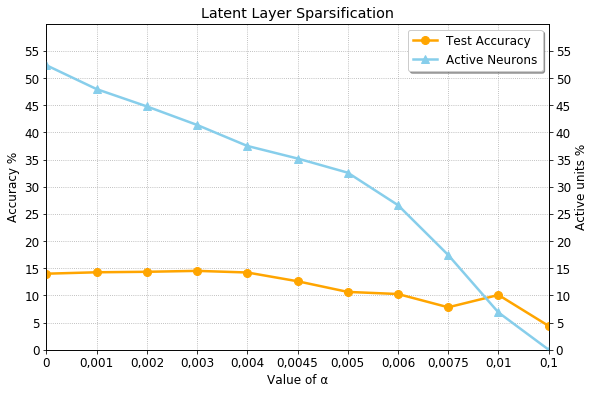}
\caption{Sparsification of \texttt{conv5\_4/dw} activations for different values of
$\alpha$ and the corresponding accuracy after the first training batch.}
\label{fig:sparsification}
\end{figure}

\subsubsection{Reducing Activations Storage in Latent Replay}
\label{subsec:active_storage}

Even if in our CORe50 case study the external storage is quite limited (e.g., 1,500 $\times$ 32KB = 48 MB for latent replay at \texttt{conv5\_4/dw}), scaling up to applications with thousands of classes could require to store much more activations and the external memory could become an issue. 
Fortunately, high layers activations can be sparsified, quantized and encoded with almost no accuracy reduction. The authors of \cite{Georgiadis2018} show that MobileNetV1 activations can be compressed up to 10 times upon proper sparsification, encoding and lossless entropy compression. In their experiments a moderate compression even leads to slightly improved accuracy because of the regularization introduced.

In the case of latent replay, we only need to sparsify the activations of the latent replay layer (and not of the entire network), potentially introducing a sort of information bottleneck. This can be easily achieved by adding an L1 term (with relative weight) to the loss function attracting toward zero the activations of the latent replay layer (see \cite{Georgiadis2018}).
We performed some preliminary experiments to sparsify activations of layer \texttt{conv5\_4/dw} during the first training batch starting from a non-sparsified ImageNet pre-trained model. Note that the weights of the latent replay layer and previous layers are frozen after the first training batch and no further sparsification can take place. The results are shown in Figure \ref{fig:sparsification}: for $\alpha=0$ (i.e., no induced sparsification) $\sim$52\% of activations are non-zero due to the natural spasification effect of the Relu activation function and the accuracy is about 14\%; as we increase $\alpha$ the amount of non-zero activation start decreasing. 
Interestingly, for $\alpha = 0.004$ we can reduce the non-zero activations from $\sim$52\% to $\sim$37\% by achieving also a slight accuracy improvement (0.22\%). By adding quantization and entropy encoding (out of the scope of this work) we believe that, analogously to \cite{Georgiadis2018}, a 10$\times$ compression is at reach with almost no accuracy loss.

\subsubsection{Comparison with Other Approaches}
\label{subsec:comparison}

\begin{figure}[h]
\centering
\includegraphics[width=\columnwidth]{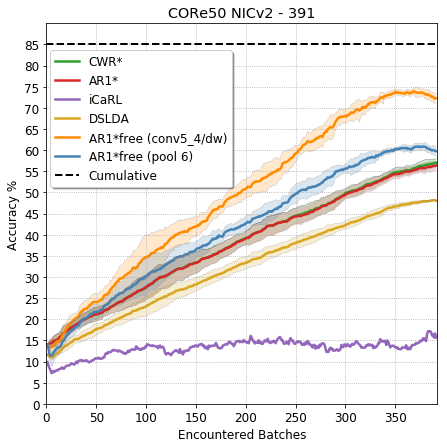}
\caption{Accuracy results on the CORe50 NICv2 – 391 benchmark of CWR*, AR1*, DSLDA, iCaRL, AR1*free (\texttt{conv5\_4}), AR1*free (\texttt{pool6}). Results are averaged across 10 runs in which the batches order is randomly shuffled. Colored areas indicate the standard deviation of each curve. As an exception, iCaRL was trained only on a single run given its extensive run time ($\sim$14 days).}
\label{fig:comparison}
\end{figure}

hile the accuracy improvement of the proposed approach w.r.t. state-of-the-art rehearsal-free techniques have been already discussed in the previous sections, a further comparison with other state-of-the-art continual learning techniques may be beneficial for better appreciating its practical impact and advantages.
In particular, while AR1* and CWR* have been already proved to be substantially better than LWF and EWC on the NICv2 - 391 benchmark, a comparison with iCaRL, one of the best know rehearsal-based technique, is worth to be considered. 

nfortunately, iCaRL was conceived for incremental class learning scenario and its porting to NIC (whose batches also include patterns of know classes) is not trivial. To avoid subjective modifications, we started from the code shared by the authors and emulated a NIC setting by: \emph{(i)} always creating new virtual classes from patterns in the coming batches; \emph{(ii)} fusing virtual classes together when evaluating accuracies. For example, let us suppose to encounter 300 patterns of class 5 in batch 2 and other 300 patterns of the same class in batch 7; while two virtual classes are created by iCaRL during training, when evaluating accuracy both classes point to the real class 5.
The hereby modified iCaRL implementation, with an external memory of 8000 patterns (much more than the 1500 used by the proposed latent replay, but in line with the settings proposed in the original paper \cite{Rebuffi2017}), was run on NICv2 - 391, but we were not able to obtain satisfactory results. In Figure \ref{fig:comparison} we report the iCaRL accuracy over time and compare it with AR1*free (\texttt{conv5\_4/dw}), AR1* (\texttt{pool6}) as well as the top three performing rehearsal-free strategies introduced before: CWR*, AR1* and DSLDA. While iCaRL exhibits better performance than LWF and EWC (as reported in \cite{Lomonaco2019}), it is far from DSLDA, CWR* and AR1*.

Furthermore, when the algorithm has to deal with a so large number of classes (including virtual ones) and training batches its efficiency becomes very low (as also reported in \cite{maltoni2019}). In Table \ref{tab:comparison} we also report the total run time (training and testing), memory overhead and accuracy difference with respect to the cumulative upper bound. We believe AR1*free (\texttt{conv5\_4/dw}) represents a good trade-off in terms of efficiency-efficacy with a limited computational-memory overhead and only at $\sim$13\% distance from the cumulative upper bound. For iCaRL the total training time was 14 days compared to a training time of less than 1 hour for the other techniques.

\subsection{Experiments on OpenLORIS}
\label{sec:openloris}



In order to show the general applicability of latent replay in different continual learning settings, we report and compare its performance also on the OpenLORIS dataset, which has been used as the main benchmark in the recent IROS 2019 \emph{``Lifelong Robotic Vision''} competition \cite{QiShe2019}. 

\begin{figure}[t]
\centering
\includegraphics[width=\columnwidth]{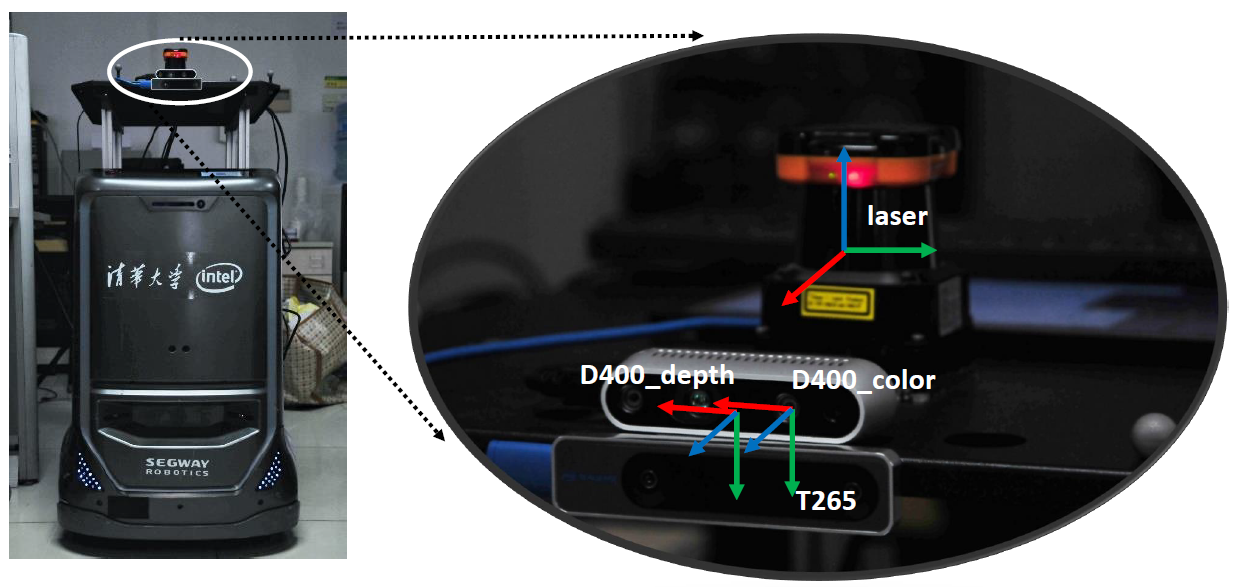}
\caption{Wheeled robot used in the \emph{Lifelong Robotic Vision} challenge at IROS 2019 \cite{QiShe2019} (left) and equipped with multiple sensors including two Real Sense cameras (right).}
\label{fig:robot}
\end{figure}

OpenLORIS is particularly interesting for continual learning in the context of robotic vision since its video sessions have been recorded on a real wheeled robot exploring its environment (see Fig. \ref{fig:robot}). In this case, however, the scenario is quite different from CORe50 NICv2 - 391: it is based on a sequence of 12 relatively large batches ($\sim$14,000 samples each) containing only new examples of the same 69 classes made available in the first batch (also known as the "New Instances" scenario, NI). For this scenario:
\begin{itemize}
\item We use a MobileNetV2 \cite{Sandler2018} pre-trained on ImageNet-1k.
\item We apply our \emph{latent replay} approach to LWF \cite{Li2016}, whose distillation steps proved to be effective to continually learn over a moderate number or large batches.

\end{itemize}

Seven finalists passed the first competition stage, and submitted their solutions to the organizers who finally produced the scoreboard reported in Table \ref{tab:challenge}. The accuracy of the proposed approach is just slightly lower that the top 1, but its inference time, replay memory and model size are significantly better. Since the challenge evaluation criteria did not include specific metrics on training efficiency, unfortunately from this experiment we cannot appreciate the training efficiency of our solution.

\begin{table}[h]
  \caption{Accuracy results at the end of the training and other metrics used for the OpenLORIS challenge benchmark.}
  \label{tab:challenge}
  \centering
  \footnotesize
  \renewcommand{\arraystretch}{1.0}
  \setlength\tabcolsep{0.07cm}
  \begin{tabular}{p{2.4cm}cccc}
    \toprule
    \multicolumn{5}{c}{OpenLORIS Challenge Results (7 finalists)}\\
    \midrule
    \thead[l]{Strategy} & \thead{Final \\Acc. \%} & \thead{Inference \\Time (s)} & \thead{Replay \\Size (Sample)} & \thead{Model \\Size (MB)}\\
    \midrule
    SDU\_BFA\_PKU & 99.56\% & 2,444.01 & 28,500 & 171.40\\
    \textbf{UniBo-Team (ours)} & \textbf{97.68\%} & \textbf{22.41} & \textbf{1,500} & \textbf{5.90}\\
    HIK\_ILG & 96.86\% & 25.42 & 0 & 16.30 \\
    Vidit98 & 96.16\% & 112.2 & 13,000 & 9.40 \\
    NTU\_LL & 93.56\% & 4,213.76 & 0 & 467.10\\
    Neverforget & 92.93\% & 89.15& 0 & 342.90\\
    Guinness & 72.9\% & 346.02 & 0 & 9.40\\
    \bottomrule
  \end{tabular}
\end{table}

\section{Real-World Deployment on Embedded Devices}
\label{sec:mobile}

\begin{figure}[h]
\centering
\includegraphics[width=\columnwidth]{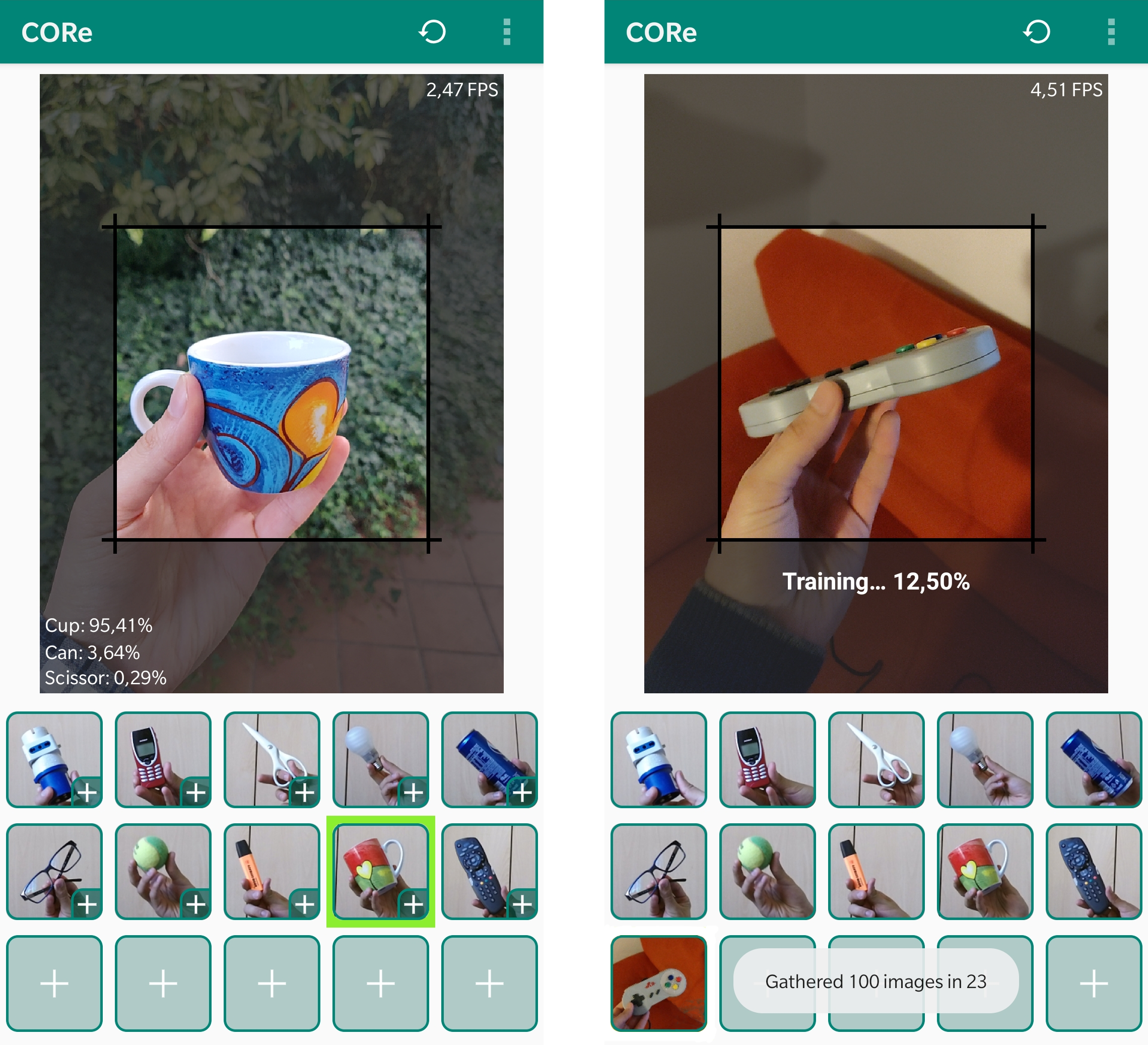}
\caption{The user interface of CORe app. The camera field of view is partially grayed to highlight the central area where the image is cropped, resized to 128 $\times$ 128 and passed to the CNN. The top three categories are returned for each image (closed set classification) and a green frame is placed around the icon of the most likely class. A training session is trigger by tapping the icon of one of the 10 existing classes or one of the (initially) five empty classes.}
\label{fig:app}
\end{figure}

The feasibility of continual learning at the edge on embedded devices is demonstrated through the development of an Android app (called CORe) for Android smartphone (see Figure \ref{fig:app}). While the app will be open-sourced and uploaded in the Google Play store upon publication of this manuscript, a video showing its functions is already available at \url{http://bit.ly/latent-replay}.

The app comes pre-trained with 10 classes (corresponding to the 10 CORe50 categories) and allows to: \emph{(i)} continually train existing classes (by learning new object/poses) and \emph{(ii)} to learn up to 5 brand new classes. As the app is launched it switches to inference mode and classifies the framed objects with an inference efficiency of about 5 fps (CPU-only with no hardware acceleration). When learning is triggered a short video of 20 seconds (at 5 fps) is acquired and the resulting 100 frames are used for continual learning that completes in less than 1 seconds after the end of the acquisition.

Behind the scenes the app is running a customized Caffe version cross-compiled for Android and using the same MobileNetV1 architecture introduced in Section \ref{sec:train_latent_replay}, here initialized to work with 15 classes. Latent replay in this case is implemented at the \texttt{pool6} layer\footnote{Placing latent reply layer at \texttt{pool6} corresponds to extending CWR* with latent replay, and leads to maximum efficiency on a CPU-only edge device.} with and external memory of 500 patterns. Low level code is written in C++ and the app interface in Java. A training session consists of 8 epochs, 5 iterations per epoch with a mini-batch size of 120 patterns: each mini-batch includes 20 original frames and 100 replay patterns. In order to speed up training during the video acquisition, a second thread immediately moves available frames forward in the CNN and caches activations at latent replay layer so that, when the acquisition is concluded, we can directly train the class specific discriminative layers. Further details and precise timing of different phases are provided in the appendix \ref{app:mobile}.

\section{Conclusions}
\label{sec:conclusion}

In this paper we showed that latent replay is an efficient technique to continually learn new classes and new instances of known classes even from small and non i.i.d. batches. State-of-the-art CL approaches such as AR1*, extended with latent replay, are able to learn efficiently and, at the same time, the achieved accuracy is not far from the cumulative upper bound (about 5\% in some cases).
The computation-storage-accuracy trade-off can be defined according to both the target application and the available resources so that even edge devices with no GPUs can learn continually from short videos, as we proved through the development of an Android application. In the future we intend to investigate: \emph{(i)} the design of more sophisticated pattern replacing strategies for the external memory to contrast the aging effect; \emph{(ii)} replacing the external memory with a generative model trained in the loop and capable of providing pseudo activations on demand.

{\small
\bibliographystyle{ieee_fullname}
\bibliography{library}
}

\appendix

\section{Implementation and Experiments Details}
For each of the proposed strategies a test accuracy curve was obtained by averaging over 5 different runs. We followed the same experimental setup as in \cite{Lomonaco2019} so that each run differs from the others by the order of the encountered batches.

Our experiments were executed in a "Ubuntu 16.04" Docker environment with a customized version of the Caffe \cite{Jia2014} framework using a single GPU. See table \ref{table:hardware} for more details of the host setup.

\section{Hyperparameters}
\label{app:hyper}
The hyperparameters used in our experiments are described in tables \ref{tab:hyper_core50} and \ref{table:renormpar}. Please note that:

\begin{itemize}
    \item We used the same naming scheme used in \cite{maltoni2019}.
    \item For AR1* and AR1*free we used a higher learning rate for the CWR layer, as described in \cite{Lomonaco2019}.
    \item In order to optimize the results for the two different rehearsal types (native rehearsal and latent replay) we chose two different values for the moving average update rate of the BatchReNorm layers \cite{loffe2017}. We found out that an higher value of the update rate was better suited for the latent version.
    \item Excluding the aforementioned update rate, we used the same hyperparameters for both the native and latent rehearsal-based experiments.
\end{itemize}

\section{Model Architecture and Memory Trade-off}
In order to assess the trade-off between accuracy, computation and memory usage we ran AR1* free using different latent replay layers. Table \ref{tab:model_arch} shows the details of the model we used, which is based on the MobileNetV1 \cite{Howard2012} with the only difference that Batch Norm layers have been replaced with Batch ReNorm ones. Here we report the network architecture as well as the pattern size and the number of weights per layer.

\section{Android Application Setup and Performance}
\label{app:mobile}
The CORe Android application has been tested on a \emph{OnePlus 6} smartphone without additional accelerators. Table \ref{table:smartphone} shows the details about this hardware platform.

In table \ref{table:profiling} we report the overall time taken for the different inference and training steps as well as the detected peak RAM usage. Note that step times, CPU usage and memory consumption may vary greatly depending on the hardware, operative system and other processes running in background.

In our experiments we used a customized version of Caffe compiled for arm64-v8a platform using OpenBLAS as the BLAS library. Additional information about the used libraries can be found in Table \ref{table:libraries}.


\begin{table}[!htb]
\centering
    \caption{Hyperparameter values used in our experiments. The selection was performed on run 0, and hyperparameters were then fixed for runs $1, 2, 3, 4$. As an exception for the long running time (around 14 days), iCaRL was trained only on run 0.}
  	\label{tab:hyper_core50}
  \begin{tabular}[H]{ll}
    \toprule
    \multicolumn{2}{c}{\textbf{CWR*}}   \\
    \cmidrule(r){1-2}
    \multicolumn{1}{c}{\textit{Parameters}}    & \multicolumn{1}{c}{\textit{MobileNet V1}}     \\
    \midrule
    Head 							& Maximal  			     \\
    $B_1$: epochs, $\eta$ (learn. rate)   		& 4, 0.001		 	      \\
    $B_i, i>1$: epochs, $\eta$ (learn. rate)   	& 4, 0.003      \\
    \toprule
    \multicolumn{2}{c}{\textbf{AR1*}}   \\
    \cmidrule(r){1-2}
    \multicolumn{1}{c}{\textit{Parameters}}    & \multicolumn{1}{c}{\textit{MobileNet V1}}     \\
    \midrule
    Head 							& Maximal  			     \\
     			     
    $w_1, w_i (i > 1)$ 							& 0.5, 0.5  	     \\
    $max_F$										& 0.001 			  \\
    
    $B_1$: epochs, $\eta$ (learn. rate)   		& 4, 0.001		 	      \\
    $B_i, i>1$: epochs   	& 4 \\
    \ \ \ \ \ $\eta$ (learn. rate, CWR layer)   	&  0.003  
    \\
    \ \ \ \ \ $\eta$ (learn. rate,other layers)   	&  0.0003 \\
    \toprule
    \multicolumn{2}{c}{\textbf{AR1* free}}   \\
    \cmidrule(r){1-2}
    \multicolumn{1}{c}{\textit{Parameters}}    & \multicolumn{1}{c}{\textit{MobileNet V1}}     \\
    \midrule
    Head 							& Maximal  			     \\
    $B_1$: epochs, $\eta$ (learn. rate)   		& 4, 0.001		 	      \\
    $B_i, i>1$: epochs   	& 4 \\
    \ \ \ \ \ $\eta$ (learn. rate, CWR layer)   	&  0.003  
    \\
    \ \ \ \ \ $\eta$ (learn. rate,other layers)   	&  0.0003 \\
    \toprule
    \multicolumn{2}{c}{\textbf{DSLDA}}   \\
    \cmidrule(r){1-2}
    \multicolumn{1}{c}{\textit{Parameters}}    & \multicolumn{1}{c}{\textit{MobileNet V1}}     \\
    \midrule
    Shrinkage   		& 1e-4		 	      \\
    $\Sigma$   	& plastic \\
    \toprule
    \multicolumn{2}{c}{\textbf{iCaRL}}   \\
    \cmidrule(r){1-2}
    \multicolumn{1}{c}{\textit{Parameters}}    & \multicolumn{1}{c}{\textit{MobileNet V1}}     \\
    \midrule
    $B_1$: epochs, $\eta$ (learn. rate)   		& 40, 0.01		 	      \\
    $B_i, i>1$: epochs, $\eta$ (learn. rate)   	& 4, 0.001      \\
    $K$   	& 8000      \\
    \bottomrule
  \end{tabular}
\end{table}

\begin{table}[H]
\caption{Experimental setup}
\centering
\begin{tabular}{lrrr}
\toprule
                     \multicolumn{1}{c}{\textit{Component}}      & \multicolumn{1}{c}{\textit{Model/Version}} \\ \midrule
 Operating System                         & Debian 8.3                         \\
 Docker                         & 18.06.1                         \\
 Nvidia Driver                         & 430.40 (CUDA 9.0, CuDNN 7)                         \\
 CPU                          & Intel(R) Xeon(R) CPU E5-2650                          \\ 
 GPU                          & GTX 1080 Ti (11 GB VRAM)                          \\ 
 RAM                       & 64 GB DDR3 (1600 MHz)                       \\ \bottomrule
\end{tabular}
\label{table:hardware}
\end{table}

\begin{table}
\caption{The architecture of model used in our experiments with neurons, weights and ops for each layer. Those information, along with the results reported in table 1, can be used to identify the most appropriate trade-off between accuracy, computation and used memory depending on the problem context.}
\label{tab:model_arch}
\begin{tabular}[H]{lrrr}
\toprule
\emph{Layer}                & \emph{Neurons} & \emph{Ops}      & \emph{Weights} \\
\midrule
Images               & 49152        & -            & -       \\
conv1                & 131072       & 3670016     		& 896     \\
conv2\_1/dw          & 131072       & 1310720       & 320     \\
conv2\_1/sep         & 262144       & 8650752       & 2112    \\
conv2\_2/dw          & 65536        & 655360        & 640     \\
conv2\_2/sep         & 131072       & 8519680       & 8320    \\
conv3\_1/dw          & 131072       & 1310720       & 1280    \\
conv3\_1/sep         & 131072       & 16908288       & 16512   \\
conv3\_2/dw          & 32768        & 327680        & 1280    \\
conv3\_2/sep         & 65536        & 8454144        & 33024   \\
conv4\_1/dw          & 65536        & 655360        & 2560    \\
conv4\_1/sep         & 65536        & 16842752        & 65792   \\
conv4\_2/dw          & 16384        & 163840        & 2560    \\
conv4\_2/sep         & 32768        & 8421376        & 131584  \\
conv5\_1/dw          & 32768        & 327680        & 5120    \\
conv5\_1/sep         & 32768        & 16809984        & 262656  \\
conv5\_2/dw          & 32768        & 327680        & 5120    \\
conv5\_2/sep         & 32768        & 16809984        & 262656  \\
conv5\_3/dw          & 32768        & 327680        & 5120    \\
conv5\_3/sep         & 32768        & 16809984        & 262656  \\
conv5\_4/dw          & 32768        & 327680        & 5120    \\
conv5\_4/sep         & 32768        & 16809984        & 262656  \\
conv5\_5/dw          & 32768        & 327680        & 5120    \\
conv5\_5/sep         & 32768        & 16809984        & 262656  \\
conv5\_6/dw          & 8192         & 81920         & 5120    \\
conv5\_6/sep         & 16384        & 8404992        & 525312  \\
conv6/dw             & 16384        & 163840        & 10240   \\
conv6/sep            & 16384        & 16793600        & 1049600 \\
pool6                & 1024         & 16384         & 0       \\
fc7                  & 50            & 51250           & 51250   \\ 
\midrule
Total                &              & 187,09M        & 3,35M  \\
\bottomrule
\end{tabular}
\end{table}


\begin{table}[H]
\caption{Batch ReNormalization parameters. The reported parameters were used in our experiments on the NICv2-391 scenario involving the CWR*, AR1* and AR1* free algorithms.}
\centering
\begin{tabular}{p{2cm}p{0.5cm}p{0.5cm}}
\toprule
                     \multicolumn{1}{c}{\textit{Parameters}}     & \multicolumn{1}{c}{\textit{Latent Replay}} & \multicolumn{1}{c}{\textit{Native Rehearsal}} \\ \midrule
$R_{max}$                       & 1.25                        & 1.25                                                 \\ 
$D_{max}$                       & 0.5                         & 0.5                             \\ 
Moving Avg. update rate &  0.99995 & 0.9999                            \\ \bottomrule
\end{tabular}
\label{table:renormpar}
\end{table}

\begin{table}[H]
\caption{The reference platform for the CORe App. In our tests we used a OnePlus 6 smartphone.}
\centering
\begin{tabular}{lp{4cm}}
\toprule
                     \multicolumn{1}{c}{\textit{Component}}      & \multicolumn{1}{c}{\textit{Model/Version}} \\ \midrule
 Model                & OnePlus 6 (A6000) \\
 Release date         & 2018, May      \\ 
 Operating System     & Android 9 (OxygenOS 9.0.9)      \\
 Chipset              & Qualcomm SDM845 Snapdragon 845  \\
 CPU                  & Octa-core (4x2.8 GHz) \\
 RAM                  & 8 GB LPDDR4X, 1866 MHz                     \\
  \bottomrule
\end{tabular}
\label{table:smartphone}
\end{table}

\begin{table}[h]
\caption{The libraries used in our Android application. In our experiments and in the Android application we use the BVLC Caffe distribution with a custom BatchReNorm layer and extended pyCaffe bindings. We also report the tools used in the build process. Note that OpenBLAS was compiled with OpenMP support (provided in the Android Ndk).}
\centering
\begin{tabular}{lr}
\toprule
    \multicolumn{2}{c}{\textbf{Used libraries}}   \\
    \cmidrule(r){1-2}
                     \multicolumn{1}{c}{\textit{Library}}      & \multicolumn{1}{c}{\textit{Version}} \\ \midrule
 Caffe (BVLC)         & Mar, 2 \\
 OpenBLAS             & 0.3.6 \\
 OpenMP               & 5.0.20140926 \\
 OpenCV               & 4.1.1      \\ 
 Boost                & 1.56.0      \\
 Gflags               & 2.2.0  \\
 Glog                 & 0.3.5 \\
 LevelDB              & 1.21.0                     \\
 Protobuf             & 3.6.1 \\
 Snappy               & 1.1.7 \\
\toprule
    \multicolumn{2}{c}{\textbf{Build tools}}   \\
    \cmidrule(r){1-2}
 \multicolumn{1}{c}{\textit{Tool}}      & \multicolumn{1}{c}{\textit{Version}} \\ \midrule
 Android Studio & 3.5.2 \\
 Gradle Android Plugin & 3.5.0 \\
 Android Ndk & 20.0.5594570 \\
 Android Clang & 8.0.7 \\
 Cmake & 3.10.2 \\
 Android Build tools & 28.0.3 \\
 
  \bottomrule
\end{tabular}
\label{table:libraries}
\end{table}

\clearpage
\makeatletter
\setlength{\@fptop}{0pt}
\makeatother

\begin{table*}[!htb]
\caption{The profiling information obtained while running the CORe App on the reference platform. The training times here reported were obtained by averaging the time taken from 5 incremental training sessions.}
\centering
\begin{tabular}{lrr}
\toprule
    \multicolumn{3}{c}{\textbf{Inference and training times (per step)}}   \\
    \cmidrule(r){1-3}
                     \multicolumn{1}{c}{\textit{Step name}}      & \multicolumn{2}{c}{\textit{Average time (ms)}} \\ \midrule
 Inference  & \multicolumn{2}{r}{255.1} \\
 Features pre-extraction         & \multicolumn{2}{r}{202.3 ms (for each pattern)} \\
 Misc. training preparation             & \multicolumn{2}{r}{1.6 (overall)} \\
 Data feeding (at latent layer)               & \multicolumn{2}{r}{64.4 (8.05 per epoch)} \\
 Forward               & \multicolumn{2}{r}{292.4 (36.55 for each epoch)}      \\ 
 Backward                & \multicolumn{2}{r}{43.6 (5.45 for each epoch)}      \\
 Update               & \multicolumn{2}{r}{1269.0 (158.63 for each epoch)} \\
 Consolidation (CWR*)        & \multicolumn{2}{r}{1.0 (overall)} \\
\toprule
    \multicolumn{3}{c}{\textbf{CPU and RAM usage}}   \\
    \cmidrule(r){1-3}
 \multicolumn{1}{c}{\textit{Phase}}      &
 \multicolumn{1}{c}{\textit{CPU usage}}      & \multicolumn{1}{c}{\textit{RAM usage}} \\ \midrule
 Inference & 17\% & 225 MB \\
 Image gathering (and feature pre-extraction) & 25\% & 240 MB  \\
 Training & 18\% & 260 MB \\
  \bottomrule
  \vspace{20cm}
\end{tabular}
\label{table:profiling}
\end{table*}

\end{document}